\documentclass[journal]{IEEEtai}

\usepackage{graphicx}
\usepackage{fancybox}
\usepackage{amssymb}
\usepackage{amsmath}
\usepackage{latexsym}
\usepackage{url}
\usepackage{xcolor}
\definecolor{newcolor}{rgb}{.8,.349,.1}
\usepackage{multirow}
\usepackage[colorlinks,urlcolor=blue,linkcolor=blue,citecolor=blue]{hyperref}

\usepackage{color,array}

\usepackage{graphicx}
\usepackage{xcolor}

\begin{document}

\title{Lifelong Learning Using a Dynamically Growing Tree of Sub-networks for Domain Generalization in Video Object Segmentation}

\author {Islam~Osman, and Mohamed~S.~Shehata,~\IEEEmembership{Member,~IEEE,}

\thanks{This paragraph will include the Associate Editor who handled your paper.}}

\maketitle

\begin{abstract}
Current state-of-the-art video object segmentation models have achieved great success using supervised learning with massive labeled training datasets. However, these models are trained using a single source domain and evaluated using videos sampled from the same source domain. When these models are evaluated using videos sampled from a different target domain, their performance degrades significantly due to poor domain generalization, i.e., their inability to learn from multi-domain sources simultaneously using traditional supervised learning. 
In this paper, We propose a dynamically growing tree of sub-networks (DGT) to learn effectively from multi-domain sources. DGT uses a novel lifelong learning technique that allows the model to continuously and effectively learn from new domains without forgetting the previously learned domains. Hence, the model can generalize to out-of-domain videos. The proposed work is evaluated using single-source in-domain (traditional video object segmentation), multi-source in-domain, and multi-source out-of-domain video object segmentation.
The results of DGT show a single source in-domain performance gain of $0.2\%$ and $3.5\%$ on the DAVIS16 and DAVIS17 datasets, respectively. However, when DGT is evaluated using in-domain multi-sources, the results show superior performance compared to state-of-the-art video object segmentation and other life
long learning techniques with an average performance increase in the F-score of $6.9\%$ with minimal catastrophic forgetting. Finally, in the out-of-domain experiment, the performance of DGT is $2.7\%$ and $4\%$  better than state-of-the-art in 1 and 5-shots, respectively. 

\end{abstract}


\section{Introduction}
\label{sec1}
\IEEEPARstart{V}{ideo} Object Segmentation (VOS) is the core of many computer vision applications \cite{a4,a27,a31,2swin}. The most common scenario in VOS is semi-supervised learning, where an initial (reference) frame annotation is provided, and the challenge is to accurately segment objects in subsequent frames in real-time, online, and with minimal memory usage for long videos. This VOS scenario is defined as in-domain (i.e., testing videos are sampled from the same distribution as training videos) 1-shot learning. Recent video object segmentation models have achieved great success, surpassing human-level performance in some cases \cite{1nserc}, using supervised learning with the aid of massive labeled training datasets. However, the performance of these models is still limited in the scope of problems they can solve and need to "increase their out-of-domain (OOD) robustness" \cite{2nserc}. In other words, they perform well on the tasks in a specific domain but are often brittle outside of that narrow domain. This shift in domain causes the performance of existing models to degrade severely. A domain shift occurs when the training/source and testing/target datasets are different. This target dataset is also referred to as OOD dataset. 
When VOS models undergo a domain shift, their performance degrades due to poor domain generalization. The model needs to learn from multiple source domains to achieve domain generalization. However, existing models are designed to learn from a single source domain, and when trained using multiple sources, these models face a problem called catastrophic forgetting, which causes these models to forget the previously learned domains while learning a new domain. Additionally, the magnitude of the problem becomes more complex when the number of labels in the new domain is small, causing the performance to degrade severely.

To the best of the knowledge of the authors, this paper is the first to address the combined problems of catastrophic forgetting and out-of-domain (OOD) few-shot learning in the context of VOS. This paper proposes a lifelong learning technique using a dynamically growing tree of sub-networks (DGT) to overcome the two problems mentioned above. Hence, DGT achieves high domain generalization. DGT has two main phases 1) the Building base knowledge phase and 2) the lifelong learning phase.  
In the first phase, the initial tree of sub-networks is built. Starting with the root node as a network trained to segment objects from all videos in the dataset (traditional VOS). Then, the videos are grouped semantically into more domain-specific groups. A child node is instantiated from the root to learn more task-specific features from each group. The process of grouping videos into more domain-specific groups is repeated until the tree is saturated (i.e., splitting the groups any further will decrease the performance due to over-fitting).
In the second phase, an agent requests a suitable network from DGT for each new video. Then, DGT sends the agent a suitable task-specific network that can converge to the video using a few examples and training iterations. After that, the agent trains the task-specific network locally and sends the trained network back to DGT. Finally, DGT decides whether to create a new node for the received network or merge it with the existing node. This process is shown in Fig.~\ref{fig:test}

To summarize, this paper proposes a Dynamically Growing Tree of sub-networks (DGT) that uses a lifelong learning technique that can learn from multiple domains, allowing the model to generalize to new OOD videos using a few labels. The contributions of this paper are:

\begin{figure}
\centering
\includegraphics[width=8.5cm]{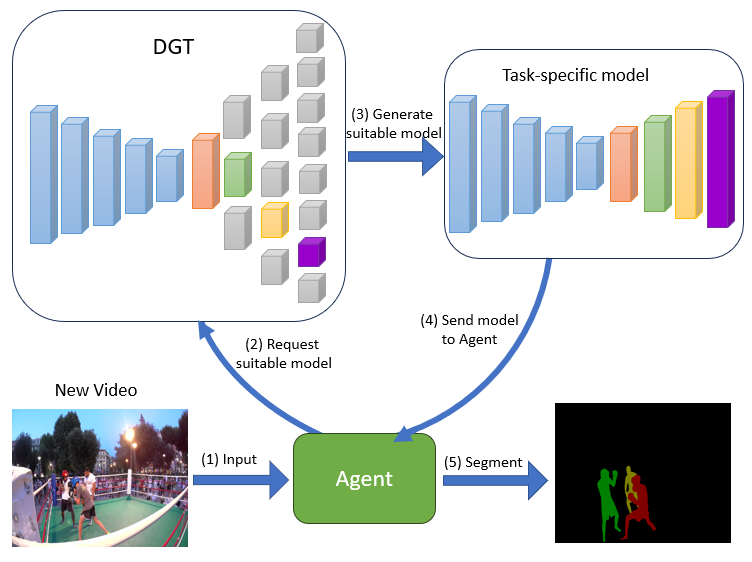}
\caption{Process of DGT in the testing phase. First, an agent requests a suitable network from DGT for the new video using a labeled reference frame. Then, DGT generates a task-specific network by selecting a suitable node for each layer of the network. Finally, the generated network is sent to the agent to segment the new video frames.}
\label{fig:test}
\end{figure}

\begin{itemize}
\item Proposing a novel lifelong learning technique to the VOS model to learn effectively from multiple source domains to achieve domain generalization.
\item Proposing a novel architecture that can dynamically increase its size to learn new tasks without forgetting previously learned tasks.
\item Achieving state-of-the-art performance in both in-domain and out-of-domain VOS.
\end{itemize}

The rest of the paper is organized as follows: Section 2 provides a literature review of existing video object segmentation models and lifelong learning techniques. The proposed work is discussed in Section 3. Section 4 depicts the experiments showing the performance of the proposed work against other state-of-the-art video object segmentation models and existing lifelong learning techniques. Finally, Section 5 concludes the paper and provides future directions.

\section{Related work}
\textbf{Video object segmentation} has witnessed many improvements due to the recent advancements in deep neural networks. Many deep learning-based video object segmentation models have been introduced and achieved promising results compared to the statistical-based models. The statistical-based models aim to create a model that represents the background using pixel intensity and/or motion vectors such as optical flow \cite{6rfnt}. Then, the modeled background is used to detect foreground objects \cite{2rfnt,37rfnt,11rfnt,10rfnt}. However, these models struggle to represent the background when environmental challenges exist in the video, such as illumination change, dynamic background, and moving camera. Deep learning models overcome these challenges. One of the early attempts in video object segmentation using deep learning models is FgSegNet \cite{27rfnt}. The authors in FgSegNet introduced a convolutional neural network with a feature pooling module to extract multi-scale features from the input image.  Another approach in video object segmentation is correspondence learning models. An example of correspondence learning models is CFBI \cite{7swin}, which matches the foreground pixels of the current frame with that of the previous frame using a global and local matching mechanism. Recently, memory networks \cite{1mskprp} have shown a great performance in video object segmentation. In memory networks, the query, key, and value are used to find a solution for current input from other inputs. In STM \cite{3swin}, the authors achieved significant success by storing and reading features of the foreground object that are predicted in the previous frame using a memory network. Another memory-network model is STCN \cite{res11}. In this model, instead of building an affinity matrix for each object in a specific memory bank, they build a single affinity matrix to learn all object relations. These models have a great performance but are limited to working on a specific domain (i.e., specific dataset). When these models are exposed to a different domain, their performance degrades due to the appearance of catastrophic forgetting and a lack of labeled data problems. The problem of catastrophic forgetting can be solved by using lifelong (continual) learning.  

\textbf{Lifelong learning} has gained much attention in the last few years. Lifelong learning aims to allow a deep learning model to learn new tasks while balancing between stability and plasticity. Model stability in lifelong learning means the ability of the model to learn new tasks without forgetting previous tasks (i.e., minimizing catastrophic forgetting). On the other hand, Model plasticity is the ability of the model to learn new tasks effectively. Usually, improving stability leads to degradation in plasticity and vice versa. Many approaches have been proposed to avoid or mitigate the problem of catastrophic forgetting \cite{13rfnt}. In LwF \cite{18pcknet}, knowledge distillation loss is used as a regularizer to enforce the predictions of the previously learned tasks to be similar to the network with the current task. Transfer without Forgetting (TwF) \cite{boschini2022transfer} uses a pre-trained network and continuously transfers knowledge from the source domain to the target domain using a layer-wise loss term. iCaRL \cite{5cl} stores samples from previous tasks and uses both stored samples and samples from the new task to update the network parameters. Exemplar-based class-incremental learning (CIL) \cite{luo2023class} is an improvement of iCaRL that increases the number of stored samples without increasing the required memory. This is done by compressing the exemplars by downsampling non-discriminative pixels and saving the compressed exemplars in the memory instead of the actual image.
PackNet \cite{1pcknet} divides the network parameters into isolated groups of parameters and assigns one of the groups of parameters to each task. All these learning techniques were introduced in the context of the image classification problem. Another approach is Model Zoo \cite{ramesh2021model}. This approach is the closest one to the proposed method. In the approach, the model expands a sub-network to learn a new task with some stored examples from old tasks. After that, all sub-networks are ensembled for prediction. The two main issues with this approach are: 1) the model size does not scale well with a large number of training tasks, as a sub-network is created for each new task. 2) the model during inference is the same as the model during training. Hence, the model inference time is high. The proposed model generates sub-networks for each group of videos that share semantic features. Hence, it scales better than Model Zoo with a large number of videos. Also, during inference, a small network is generated from the tree of sub-networks, which allows the inference time of the model to be very low.

\section{The proposed model: Dynamically Growing Tree (DGT)}

\subsection{Architecture}
\label{arch}
The proposed model architecture comprises two main components: a deep-learning network and a tree structure that holds different values for the network parameters. The parameters' value stored in each tree node can replace parts of the network parameters to solve a specific task (i.e., segment one or more videos).

\begin{figure}
\centering
\includegraphics[width=0.45\textwidth]{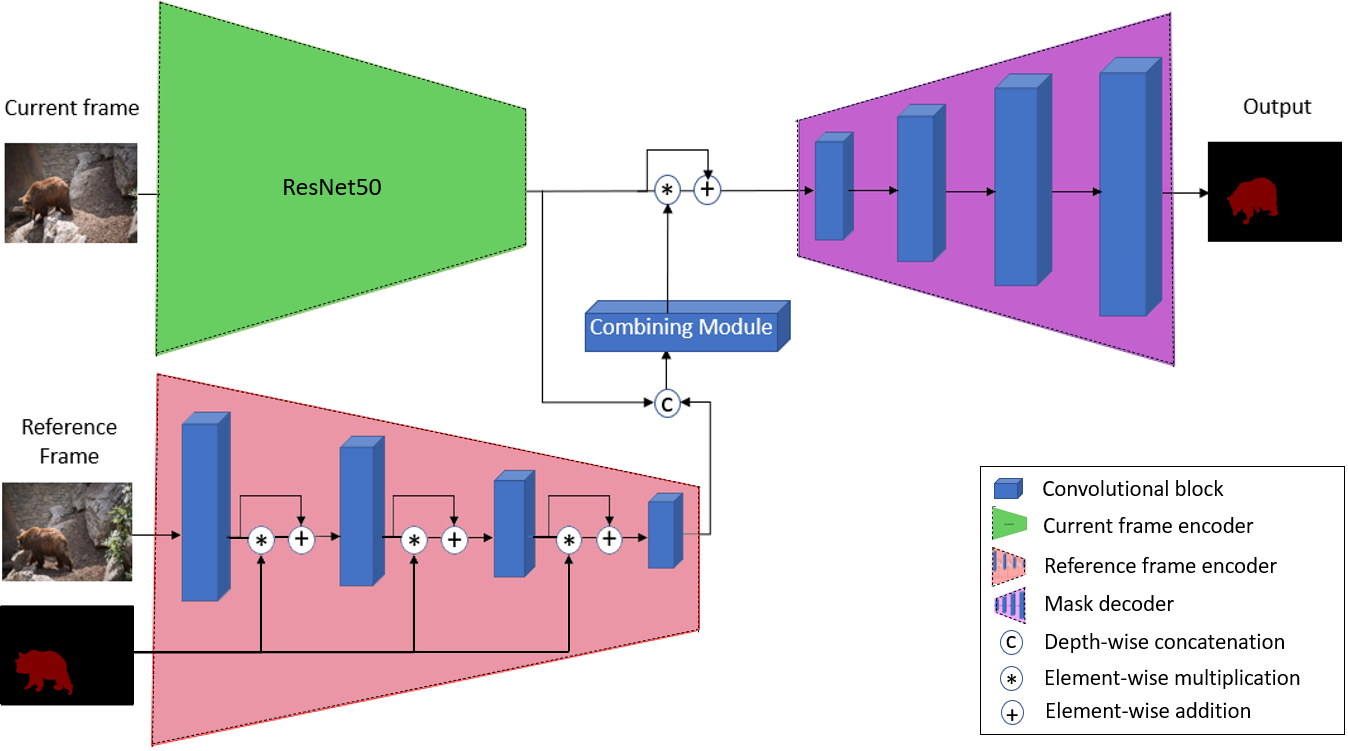}
\caption{Architecture of the proposed network.}
\label{fig:net}
\end{figure}

\textbf{The network architecture:} The architecture consists of a current frame encoder, reference frame encoder, combining module, and mask decoder. The current frame encoder is ResNet50 \cite{rsnt} pre-trained using ImageNet \cite{41rfnt}. The reference frame encoder consists of three blocks. Each block has $2$ convolutional layers followed by a max-pooling layer. Between each of the two blocks, the output feature map is multiplied by the ground truth and then added to the original output feature map to highlight the feature of the target objects only. The combining module first concatenates the output feature map from the current and reference frame encoders and then processes both of them using two convolutional layers. Then, the output is multiplied by the current frame feature map, and the result is added to the current frame feature map. This is done to highlight the features of the target objects in the current frame. Finally, the mask decoder uses the highlighted feature map of the current frame as an input. The mask decoder consists of $4$ blocks. Each block has $2$ convolutional layer followed by an up-sampling layer. After the $4$ blocks, one convolutional layer with one kernel of size $1\times1$ and sigmoid activation function is used to produce the output mask. The detailed architecture is shown in Fig.~\ref{fig:net}.

\textbf{The tree-structure:} It consists of tree nodes. Each tree node has a list of videos assigned to it, a list of children nodes, a link to the parent node, and, most importantly, the list of parameters that can be used to segment objects from all videos in the video list of this node. The size of the parameters list in the node depends on the node depth. If a node is at depth zero (root node), the list of parameters will have all $4$ decoder block parameters. If a node is at depth 1, the list of parameters will have $3$ decoder block parameters. The number of decoder blocks to be stored decreases as the node depth increases. The point of decreasing the number of parameters in deeper nodes is to decrease the overall size of stored parameters in the tree without affecting the performance because deeper nodes have fewer videos assigned to them and, hence, require fewer parameters to segment them. As shown in Fig.~\ref{fig:net}, the tree is only applied for the decoder parameters. While the encoders and combining module parameters are shared between all videos. 

\textbf{Generating a task-specific network from the tree:} Generating a network from any node in the tree will change only the decoder parameters of the network. Assume that the suitable node in the tree for segmenting a given video is at depth $d=3$. Let $G=\{\phi_0,\phi^i_1,\phi^j_2,\phi^k_3\}$ be the list of nodes in the path from the root node to the selected node, where $\phi_0$ is the root node, $\phi^i_1$ is the $i^{th}$ node in depth $d=1$, and so on. First, replace all four decoder block parameters with parameters stored in $\phi_0$. Then, replace the last three decoder blocks with parameters from $\phi^i_1$. After that, replace the last two decoder blocks parameters from $\phi^j_2$. Finally, the last decoder parameters are replaced with parameters from $\phi^k_3$. From this example, we can see that the maximum depth of the tree is $L-1$, where $L$ is the number of decoder blocks. In Fig.~\ref{fig:test}, the color of the decoder block shows which node in the tree is used to replace the parameters of each decoder block.

\subsection{The proposed lifelong learning procedure}
\label{llp}
The training of DGT is done based on a novel lifelong learning procedure and is divided into two phases: 1) building the base DGT and 2) growing the base DGT.

\textbf{Building the base DGT:}
The process of building the base DGT has two main steps: 1) pre-training and 2) repeated sequential training. 
The pre-training step: The network is trained using all videos in the dataset. In this step, all network parameters are updated. Then, the decoder parameters are stored in the tree's root node. 

The repeated sequential training step: In this step, the proposed model has access to only 1 video at a time and no access to previous or next videos (lifelong-learning configuration). For the first video, a new child is instantiated from the root node (i.e., a copy of the last $L-d$ decoder blocks parameters from the root node, where $L$ is the number of decoder blocks, and $d$ is the depth of the current node). The decoder parameters of the network are replaced by the root parameters. Then, all network parameters are frozen (not to be updated) except the parameters of the last $L-d$ decoder blocks. The trainable parameters of the network are updated using the first video. After that, the updated parameters are stored in the child node. For each other video, multiple instances of the network are generated; one is generated using the parameters of the root node, and one is generated for each child node. Then, the $\mathcal{F}$-score is calculated from the output of all network instances using the first frame of the given video. If the network instance with the highest $\mathcal{F}$-score is the one generated from the root node, a new child node is created for the given video. Otherwise, the video is assigned to the child node with the highest $\mathcal{F}$-score. 

Since this child node already exists (i.e., other videos are assigned to it), updating the parameters of this node will affect the performance of the previous videos due to catastrophic forgetting. To mitigate the problem of catastrophic forgetting,  the parameters update is weighted by a factor such that the parameters that are sensitive to changes (i.e., important to previous videos) have a small weight, and other parameters have higher weights. The parameter sensitivity is calculated using the fisher information matrix \cite{fim}. The Fisher information matrix measures how much a change in the parameter's value would change the network's prediction and is calculated as follows:

\begin{equation}
    E_xD_{KL}(P_\phi(y|x) || P_{\phi+\delta}(y|x)) = \delta^T F_\phi\delta + O(\delta^3)
\end{equation}
where $\delta$ is a small perturbation vector with the size $|\phi|$, and $F_\phi$ is the fisher information matrix with size $|\phi|\times|\phi|$. Since we need to calculate $F_\phi$ in each iteration, we approximate $F_\phi$ as a diagonal matrix to be easily computed as follows:

\begin{equation}
    \hat{F}_\phi = \frac{1}{N}\overset{N}{\underset{i=1}{\sum}} E_{y\sim P_\phi(y|x_i)}(\nabla_\phi log P_\phi(y|x_i))^2
\end{equation}
Then, $\hat{F}_\phi$ is normalized to be $\in [0,1]$. If a parameter has the value of $0$ in $\hat{F}_\phi$, this parameter has no effect on the network's output (not an important parameter) and vice versa. Finally, each parameter update is weighted by its inverted $\hat{F}_\phi$:
\begin{equation}
\label{eq:update}
\phi = \phi - \lambda (1-\hat{F}_\phi) \nabla_{\phi} \ell(\hat{y},y)
\end{equation}
By doing this, updating parameters that are important to previous videos will be minimal as the corresponding value of $(1-\hat{F}_\phi)$ is close to $0$. On the other hand, the parameters that are not important will be updated normally as their corresponding value of $(1-\hat{F}_\phi)$ is close to $1$. The $\lambda$ is the learning rate.
After updating the network parameters, if the network's performance on a video after being assigned to a child node is lower than the performance of the network generated from the root node. The video is removed from the child node. This degradation in performance is caused due to either a lack of network plasticity due to weighting the parameters update with the fisher information matrix or the network becomes very task-specific that it over-fits on the training frames of the video.

After assigning each group of videos to a child node, each child node is considered a root node of the sub-tree, and the process of clustering videos (sequential training) is repeated until reaching a stopping condition. The stopping conditions are: 1) the root node of the sub-tree has only 1 video. 2) reaching the maximum depth of the tree. 3) the average performance of the child node is lower than the average performance of the root node. 
The result of this phase (building the base DGT) is an unbalanced tree of sub-networks. The tree is unbalanced due to the different stopping conditions that may terminate some nodes' clustering process at an early depth. 

\begin{figure*}
\centering
\includegraphics[width=0.8\textwidth]{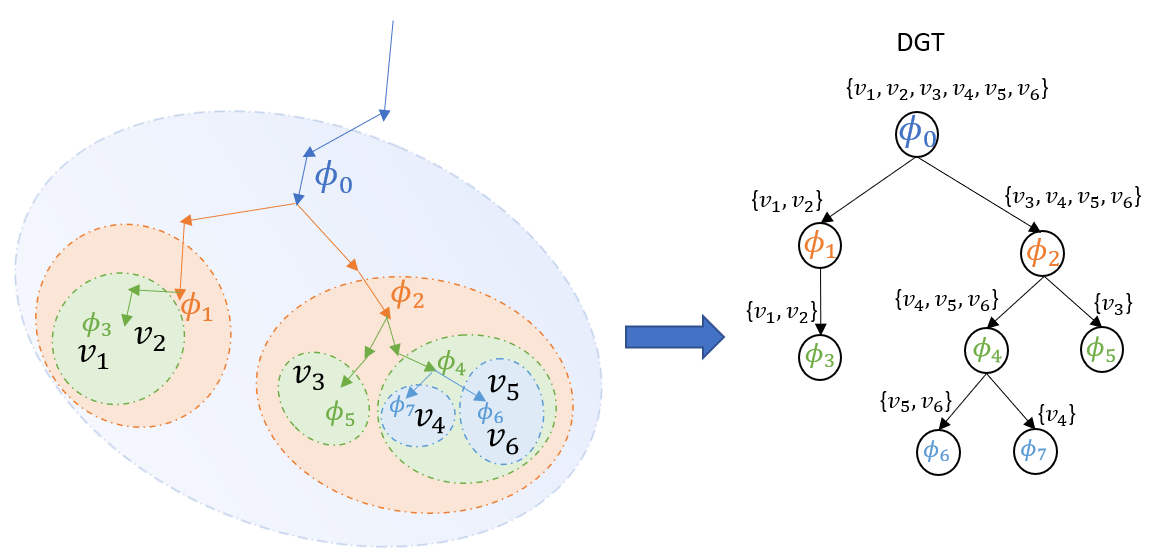}
\caption{Visualization of building a base DGT given $6$ videos and a decoder with $4$ layers. On the left side is coarse to fine clustering of videos. On the right side is the produced DGT.}
\label{fig:dgt}
\end{figure*}

An example of building a base DGT for a dataset of videos is visualized in Fig.~\ref{fig:dgt}. The blue area represents the tree initialization step where the network is trained using all videos. The orange, green, and cyan areas represent three coarse to fine clustering iterations. Each circle represents a node in the tree, such that $\phi_0$ is the set of parameters of the root node, $\phi_1$ and $\phi_2$ are parameters of nodes created at depth $d=1$, $\phi_3, \phi_4,$ and $\phi_5$ are parameters of nodes created at depth $d=2$, and so on. Since the number of decoder layers is $4$, then the maximum depth of the tree is $4$. The location of $v_i$ in the figure represents the optimal parameters to solve it, and the colored arrows represent training steps.

\textbf{Growing of DGT}:
In this phase, the base tree grows to learn new unseen OOD videos. The most suitable node in DGT for each new video is found using a greedy breadth-first search fashion. Starting from the root node as a parent node, the $\mathcal{F}$-score is calculated for each network instance generated using the parent and all children nodes. If the parent's $\mathcal{F}$-score is higher than all children, the search stops and returns the parent as the suitable node. If at least one child has an $\mathcal{F}$-score higher than the parent, the child with a maximum $\mathcal{F}$-score will act as a root to a sub-tree. This process is repeated until a parent node with a higher $\mathcal{F}$-score than all its children are found or a terminal node is reached. Finally, based on the outcome of the search algorithm, if the returned node by the search is a terminal node (i.e., a node that has no children), DGT assigns the new video to this node and updates the parameters as mentioned before in Eq.~\ref{eq:update}. On the other hand, if the node is an internal node, a new child node is instantiated as a copy of the parent node. 

\subsection{DGT video object segmentation during testing}
Once the training is complete, as described above, DGT is ready to segment objects from in-domain or out-of-domain videos. We define the agent as an interface between the dataset and the DGT. For each video in the dataset, the agent sends the reference frame of the current video to DGT and prompts it to find a suitable network to segment this frame. Then, DGT will create a task-specific network generated from a suitable node in the tree. The search for the suitable node is done by using the greedy algorithm mentioned before. The task-specific network is generated from the selected node as illustrated in Sec.~\ref{arch}. Finally, the agent will use the task-specific network generated by DGT to segment all frames of the current video. If the video has multiple objects, a mask for each object in the reference frame is passed through the network separately (i.e., if a frame has three objects, the encoder will process the input frame only once, but the decoder will process the input along with corresponding object reference frame three times, one time for each object). Then, the output masks are combined to form the multiple-object output of each frame.

\section{Experiments}

\subsection{Datasets}

\textbf{Densely Annotated VIdeo Segmentation (DAVIS) 2016 and 2017} \cite{35rfnt,13swin} is a video object segmentation dataset. Each video has a number of frames ranging from $50$ to $104$. In DAVIS16, a single object is annotated, which is the object of interest in this video. The videos are split into $33$ training videos and $20$ testing videos. DAVIS17 has multiple object annotations.

\textbf{SegTrackV2} \cite{22rfnt} is a small-scale dataset of videos, where each video has multiple foreground objects. The number of videos in the dataset is $13$, and the number of frames ranged from $21$ to $279$ in each video. The videos are split into $7$ training videos and $6$ validation videos.

\textbf{YouTube Video Object Segmentation 2018 (YT-VOS18)} \cite{11swin} is a large-scale dataset that was collected from YouTube video clips. The dataset has 94 different object categories and annotations for more than $190,000$ objects.

\textbf{ChangeDetection.Net (CDNet)} \cite{43rfnt} is a benchmark dataset used to detect changes between frames of a video. These changes are the foreground objects. Different challenges are presented in this dataset, such as illumination change, camera jitters, night vision, shadows, and dynamic background. The number of videos in each challenge ranges between $4$ to $6$ videos.

\subsection{Implementation details}
\textbf{Training.} The learning rate is $1e-5$ and the batch size is $16$. The optimizer is AdamW \cite{kingma2014adam} with a weight decay of $0.05$. The loss function is binary cross-entropy. The input image resolution is $320\times240$. Random cropping is applied on both the reference RGB image and corresponding ground truth to regularize the network. In the first phase of DGT, the root node is trained using all training videos in a given dataset for $50$ epochs. After that, the videos are hierarchically and semantically split into groups by the network, and a new tree node is created for each group. Each node is trained using the videos in the corresponding group for $10$ epochs. The stopping conditions for video splitting are: 1) node has only one video, 2) further splitting would decrease the average performance of the given group of videos, and 3) reached max depth of the tree. In the second phase, the learning rate is $1e-4$, and the number of epochs is $10$.
The network architecture details are as follows: The current frame encoder is ResNet50. The reference frame encoder has three blocks. The first block has $2$ convolutional layers with $32$ filters in each layer. In the second block, $2$ convolutional layers with $64$ filters in each layer.
The third block has $2$ convolutional layers with $128$ filters in each layer. Between every two blocks, a max-pooling layer is added, followed by a multiplication of the output feature map and the ground truth, and then the output is added to the original output feature map to highlight the
feature of the target objects only, as shown in Fig.~\ref{fig:net}. The combining module uses the concatenation of both the current frame encoder and reference frame encoder as input. The combining module comprises $2$ convolutional layers with $512$ filters in each layer. Then, the output of the combining module is multiplied by the current frame feature map, and the result is added to the current frame feature map. This is done to highlight the features of the target objects in the current frame. Finally, the mask decoder consists of $4$ blocks. Each block has two convolutional layers. The number of filters in each layer in each block is $256$, $128$, $64$, and $32$. The output layer is a convolutional layer with a single filter with dimensions $1\times1$ followed by a Sigmoid activation function. These configurations are for the large network (Net$_L$). We also report the results of a smaller network (Net$_S$) to examine the scalability of the proposed DGT in both model size and model performance using two networks with different numbers of parameters. The small version of the network is the same as the large network except for the number of filters in the decoder

\subsection{Evaluations and results}
\label{sec:eval}
To show the strength of the proposed DGT, three different experiments are used to evaluate its performance: 1) traditional video object segmentation using a single-source in-domain, 2) sequential training from multi-sources in-domain, and 3) few-shot learning from out-of-domain videos. 

In the first experiment, DGT and state-of-the-art methods are trained using a group of videos and tested using another group of unseen videos within the same domain (i.e., the same dataset). The datasets used in this experiment are DAVIS16 \cite{35rfnt}, DAVIS17\cite{13swin} and YT-VOS18 \cite{11swin}. 

In the second experiment, we address the problem of catastrophic forgetting and the inability to learn from multiple source domains. In this experiment, the results of DGT are compared with state-of-the-art VOS models and lifelong learning techniques. All models are trained using three datasets sequentially. These datasets are YT-VOS18 \cite{11swin}, CDNet \cite{35rfnt}, and DAVIS16 \cite{43rfnt}.

The last experiment shows the strength of DGT to generalize over out-of-domain videos with limited labeled data. In this experiment, the trained DGT from the previous experiment (trained using three datasets sequentially) is fine-tuned using a few labeled frames (1-shot and 5-shots) from each video in the SegTrackV2 \cite{22rfnt} dataset that is not seen during training. The results are compared with state-of-the-art VOS models and lifelong learning techniques.

Finally, an analysis of the results from the three experiments and an ablation study are presented with respect to runtime, growth of the model size, and the performance of different configurations of DGT.

\subsubsection{\textbf{Single-source in-domain video object segmentation}}

In this experiment, the datasets DAVIS16, DAVIS17, and YT-VOS18 are split video-wise into training and testing, such that videos in the testing set are different from videos in the training set. The models are trained using the training videos in single source domain video object segmentation evaluation. Then, in the testing phase, the models can access one labeled frame from each video in the testing set to fine-tune the model. For the evaluation metrics, every dataset has its own recommended metrics. For DAVIS16 and DAVIS17, the standard metrics are the mean Jaccard index $\mathcal{J}$ that is calculated as the average IoU between the prediction and the ground truth, mean boundary $\mathcal{F}$-score and their average $\mathcal{J}\&\mathcal{F}$ to evaluate segmentation accuracy. In YT-VOS18, the standard metrics are $\mathcal{J}\&\mathcal{F}$-score for both videos with seen objects and videos with unseen objects. Then, the average of all metrics is reported as $\mathcal{G}$.  
Table~\ref{tb:davis16}, \ref{tb:davis17}, and \ref{tb:yt18} show the results of the proposed DGT against state-of-the-art methods. DGT outperforms other models in two datasets, DAVIS16 and 17, using both networks, Net$_S$ and Net$_L$. The performance gain using our model is $0.2\%, 3.5\%$ in $\mathcal{J}\&\mathcal{F}$ for the datasets DAVIS16 and DAVIS17, respectively. On the other hand, DGT-Net$_L$ is slightly behind the state-of-the-art in YT-VOS18 by a small gap of $0.1\%$ in the average metric $\mathcal{G}$.
\begin{table}[]
\centering
\caption{Single-source in-domain video object segmentation results of DGT compared with state-of-the-art models on the DAVIS16 dataset.}
\begin{tabular}{l|c|c|c}
\hline
\multicolumn{1}{l|}{Model}              & \multicolumn{1}{c|}{$\mathcal{F}$}& \multicolumn{1}{c|}{$\mathcal{J}$}& \multicolumn{1}{c}{$\mathcal{F}$\&$\mathcal{J}$} \\ \hline

CINM \cite{res1}                                                                           & 0.850  & 0.834 & 0.842                                       \\
LUCID \cite{17rfnt}                                             & 0.820    & 0.839 & 0.829                      \\
STRCF \cite{20rr}                                                        & 0.816  & 0.774 & 0.795                        \\

DHS \cite{23rr}                                                        & 0.815 & 0.701 & 0.758                         \\
MiVOS \cite{res3}                                                      & 0.924  & 0.897 & 0.910                        \\

KMN \cite{res4}                                                      & 0.915   & 0.895 & 0.905                       \\

STM \cite{res5}                                                      & 0.901  & 0.887 & 0.894                        \\
RMNet \cite{res6}                                                      & 0.887    & 0.889 & 0.888                      \\
MHP-VOS\cite{res7}                                                      & 0.895    & 0.876 & 0.885                      \\
RANet+\cite{res8}                                                      & 0.876    & 0.866 & 0.871                      \\
CFBI \cite{res9}              & 0.905  & 0.883 & 0.894                        \\
CFBI+ \cite{res9}              & 0.911  & 0.887 & 0.899                        \\
HMMN \cite{res10}              & 0.920  & 0.896 & 0.908                        \\
STCN \cite{res11}              & 0.930 & 0.904 & 0.917    \\                                  
UOVOS  \cite{48rfnt}                                                & 0.772   & 0.724 & 0.748                            \\
RDE \cite{li2022recurrent}                                                   & 0.932  & 0.907 & 0.920                         \\
FgSegNet \cite{27rfnt}                                                   & 0.847  & 0.805 & 0.826                         \\
XMem \cite{cheng2022xmem}                                                      & 0.927  & 0.904 & 0.915                       \\
 REFNet-TBPI \cite{0rfnt}       & 0.927  & 0.891 & 0.909                       \\\hline
\multicolumn{1}{l|}{\textbf{DGT-Net$_S$ (ours)}}                     & 0.940 & 0.902 & 0.921
\\\multicolumn{1}{l|}{\textbf{DGT-Net$_L$ (ours)}}                     & \textbf{0.944} & \textbf{0.900} & \textbf{0.922} 
\\ \hline
\end{tabular}
\label{tb:davis16}
\end{table}

\begin{table}[]
\centering
\caption{Single-source in-domain video object segmentation results of DGT compared with state-of-the-art models on the DAVIS17 dataset.}
\begin{tabular}{l|c|c|c}
\hline
\multicolumn{1}{l|}{Model}              & \multicolumn{1}{c|}{$\mathcal{F}$}& \multicolumn{1}{c|}{$\mathcal{J}$}& \multicolumn{1}{c}{$\mathcal{F}$\&$\mathcal{J}$} \\ \hline

STM \cite{res5}                                                                           & 0.843  & 0.792 & 0.818                                       \\
HMMN \cite{17rfnt}                                             & 0.875    & 0.819 & 0.847                      \\
RPCM \cite{20rr}                                                        & 0.860  & 0.813 & 0.837                        \\
STCN \cite{21rr}                                                    & 0.886  & 0.822 & 0.854                       \\

AOT \cite{23rfnt}                                                  &   0.875       & 0.823 & 0.849                 \\

RDE \cite{23rr}                                                        & 0.875 & 0.808 & 0.842                         \\
XMem \cite{cheng2022xmem}                                                      & 0.895  & 0.829 & 0.862                       \\
DeAOT \cite{res3}                                                      & 0.882  & 0.822 & 0.852                        \\

MiVOS \cite{res4}                                                      & 0.874   & 0.817 & 0.845                       \\
ISVOS \cite{wang2023look}                                                   & 0.919  & 0.845 & 0.882                         \\
RDE \cite{li2022recurrent}                                                   & 0.900  & 0.821 & 0.861                         \\
FgSegNet \cite{27rfnt}                                                   & 0.847  & 0.805 & 0.826                         \\

\hline
\multicolumn{1}{l|}{\textbf{DGT-Net$_S$ (ours)}}                     & 0.921 & 0.849 & 0.885
\\\multicolumn{1}{l|}{\textbf{DGT-Net$_L$ (ours)}}                     & \textbf{0.932} & \textbf{0.862} & \textbf{0.897}
\\ \hline
\end{tabular}
\label{tb:davis17}
\end{table}

\begin{table}[]
\centering
\caption{Single-source in-domain video object segmentation results of DGT compared with state-of-the-art models on the YT-VOS18 dataset.}
\begin{tabular}{l|c|c|c|c|c}
\hline
\multicolumn{1}{l|}{Model}              &\multicolumn{1}{c|}{$\mathcal{G}$}& \multicolumn{1}{c|}{$\mathcal{J}_s$}& \multicolumn{1}{c|}{$\mathcal{F}_s$} & \multicolumn{1}{c|}{$\mathcal{J}_u$}& \multicolumn{1}{c}{$\mathcal{F}_u$}  \\ \hline

STM \cite{res5}                                                                           & 0.794  & 0.797 & 0.842 & 0.728 & 0.809                                       \\
HMMN \cite{17rfnt}                                             & 0.826    & 0.821 & 0.870 & 0.768 & 0.846                      \\
CFBI \cite{res9}              & 0.814  & 0.811 & 0.858 & 0.753 & 0.834                        \\                       
STCN \cite{21rr}                                                    & 0.830  & 0.819 & 0.865 & 0.779 & 0.857                       \\

AOT \cite{23rfnt}                                                  &   0.855       & 0.845 & 0.895 & 0.796 & 0.882                 \\

XMem \cite{cheng2022xmem}                                                      & \textbf{0.857}  & 0.846 & 0.893 & \textbf{0.802} & \textbf{0.887}                       \\

MiVOS \cite{res4}                                                      & 0.826   & 0.811 & 0.856 & 0.777 & 0.862                      \\

\hline
\multicolumn{1}{l|}{\textbf{DGT-Net$_S$ (ours)}}                     & 0.832 & 0.825 & 0.870 & 0.776 & 0.857
\\\multicolumn{1}{l|}{\textbf{DGT-Net$_L$ (ours)}}                     & 0.856 & \textbf{0.851} & \textbf{0.896} & 0.797 & 0.880
\\ \hline
\end{tabular}
\label{tb:yt18}
\end{table}

\subsubsection{\textbf{Sequential training from multi-sources in-domain}}
\label{msd}

This experiment highlights the proposed model's effectiveness in continuous learning with minimal forgetting compared with state-of-the-art lifelong learning techniques and video object segmentation models. The selected continual learning techniques are EWC \cite{2cl}, iCaRL \cite{5cl}, Model Zoo \cite{ramesh2021model}, TwF \cite{luo2023class}, and CIL \cite{luo2023class}. All these lifelong learning techniques are used to train our Net$_L$ to have a fair comparison with our proposed model (i.e., same network but trained using different lifelong learning techniques). Additionally, results of the state-of-the-art video object segmentation model (XMem \cite{cheng2022xmem}) are reported.
We selected three datasets with different tasks to increase the catastrophic forgetting in this experiment. These datasets are YT-VOS18 \cite{11swin}, CDNet \cite{35rfnt}, and DAVIS16 \cite{43rfnt}. These three datasets have three different tasks. YT-VOS18 is a multiple object segmentation dataset (i.e., segment specific objects while tracking them across the video), and CDNet is a moving object detection dataset (i.e., all foreground objects have to be detected and have the same label). DAVIS16 is similar to YT-VOS18 but is a single object segmentation dataset.
Each model in this experiment is trained using these datasets sequentially (i.e., one dataset at a time). The reported metrics are the $\mathcal{F}$-score and catastrophic forgetting factor $CF$. In more detail, after training on each video, the $\mathcal{F}_v$ is calculated for all previous videos, and $\mathcal{F}$ is the average of all $\mathcal{F}_v$:

\begin{equation}
    \mathcal{F}_v = \frac{1}{v}\overset{v}{\underset{i=1}{\sum}} \mathcal{F}_{v,i}
\end{equation}
\begin{equation}
    \mathcal{F} = \frac{1}{N}\overset{N}{\underset{i=1}{\sum}} \mathcal{F}_i
\end{equation}
where $\mathcal{F}_{v,i}$ is the $\mathcal{F}$-score of video $i$ after training using $v$ videos, and $N$ is the total number of videos in a dataset.
The forgetting $CF$ is calculated as the average forgetting of each video $CF_v$. The catastrophic forgetting of each video is calculated as the average degradation in $\mathcal{F}$-score for video $v$ after training using all videos in the dataset:
\begin{equation}
    CF_v = \frac{1}{v}\overset{v}{\underset{i=1}{\sum}} \mathcal{F}_{v,v}-\mathcal{F}_{i,v}
\end{equation}
\begin{equation}
    CF = \frac{1}{N}\overset{N}{\underset{i=1}{\sum}} CF_i
\end{equation}

After training using all three datasets, all models are evaluated using a testing set in all three datasets without re-training or fine-tuning. Table~\ref{tb:gfs} shows the results of this experiment. Each dataset has two columns, one for the $\mathcal{F}$-score and one for the partial catastrophic forgetting that occurred in this dataset. The last two columns are the average performance on all three datasets. The table shows that the proposed DGT significantly outperforms XMem by $15.4\%$ improvement in the average $\mathcal{F}$-score of all three datasets. Also, DGT outperforms state-of-the-art lifelong learning techniques by an average of $6.9\%$ in $\mathcal{F}$-score of all three datasets.
From this table, we observed that the memory network XMem \cite{cheng2022xmem} is good at dealing with catastrophic forgetting but performs poorly when initialized with an empty mask with no foreground objects (this is why XMem performance on CDNet is low). Additionally, using a lifelong learning technique both enhances performance and lowers catastrophic forgetting when training using multiple datasets sequentially. Finally, the proposed DGT outperforms state-of-the-art VOS models and state-of-the-art lifelong learning techniques.

\begin{table*}[]
\centering
\caption{Multi-sources in-domain video object segmentation results of DGT compared to state-of-the-art methods on all three datasets..}
\begin{tabular}{l|c|c|c|c|c|c|c|c}
\hline
 \multirow{2}*{Model}              & \multicolumn{2}{c|}{YT-VOS18}& \multicolumn{2}{c|}{CDNet}& \multicolumn{2}{c|}{DAVIS17}& \multicolumn{2}{c}{Average} \\ \cline{2-9}
 & $\mathcal{F}$& $CF$&$\mathcal{F}$&$CF$&$\mathcal{F}$&$CF$&$\mathcal{F}$&$CF$\\ \hline
XMem \cite{cheng2022xmem}                                                                           & 0.713  &0.177& 0.437 &0.037& 0.905&0.021&0.685&  0.078                                     \\

EWC-Net$_L$ \cite{2cl}                                                        & 0.635 &0.091& 0.498      &0.101& 0.855&0.049&0.662 &  0.080                       \\
iCarL-Net$_L$ \cite{5cl}                                                      & 0.660  &0.070& 0.568         &0.051& 0.870&0.033&0.699 &  0.051             \\
ModelZoo-Net$_L$ \cite{ramesh2021model}                                                      & 0.762  &0.038& 0.673               &0.029& 0.875&0.019&0.770 & 0.028         \\

TwF-Net$_L$ \cite{boschini2022transfer}                                                      & 0.690   &0.056& 0.587  &0.046& 0.827&0.031&0.701& 0.044                   \\

CIL-Net$_L$ \cite{luo2023class}                                                      & 0.715  &0.041& 0.602  &0.043& 0.830&0.034&    0.715    &  0.039              \\ \hline
\multicolumn{1}{l|}{\textbf{Net$_S$ (ours)}}                     & 0.467 &0.162& 0.479 &0.152& 0.833  &0.075&    0.593      & 0.129       \\
\multicolumn{1}{l|}{\textbf{Net$_L$ (ours)}}                     & 0.498 &0.209& 0.485 &0.234& 0.861 &0.064&     0.614 &  0.169           \\
\multicolumn{1}{l|}{\textbf{DGT-Net$_S$ (ours)}}                     & 0.827 &0.036& 0.639 &0.020& 0.905 &0.023& 0.790&0.026 
\\ 

\multicolumn{1}{l|}{\textbf{DGT-Net$_L$ (ours)}}                     & \textbf{0.860} &\textbf{0.028}& \textbf{0.733} &\textbf{0.021}& \textbf{0.926} &\textbf{0.016}& \textbf{0.839}  &\textbf{0.021}               \\\hline
\end{tabular}
\label{tb:gfs}
\end{table*}

\begin{figure}
\centering
\includegraphics[width=0.45\textwidth]{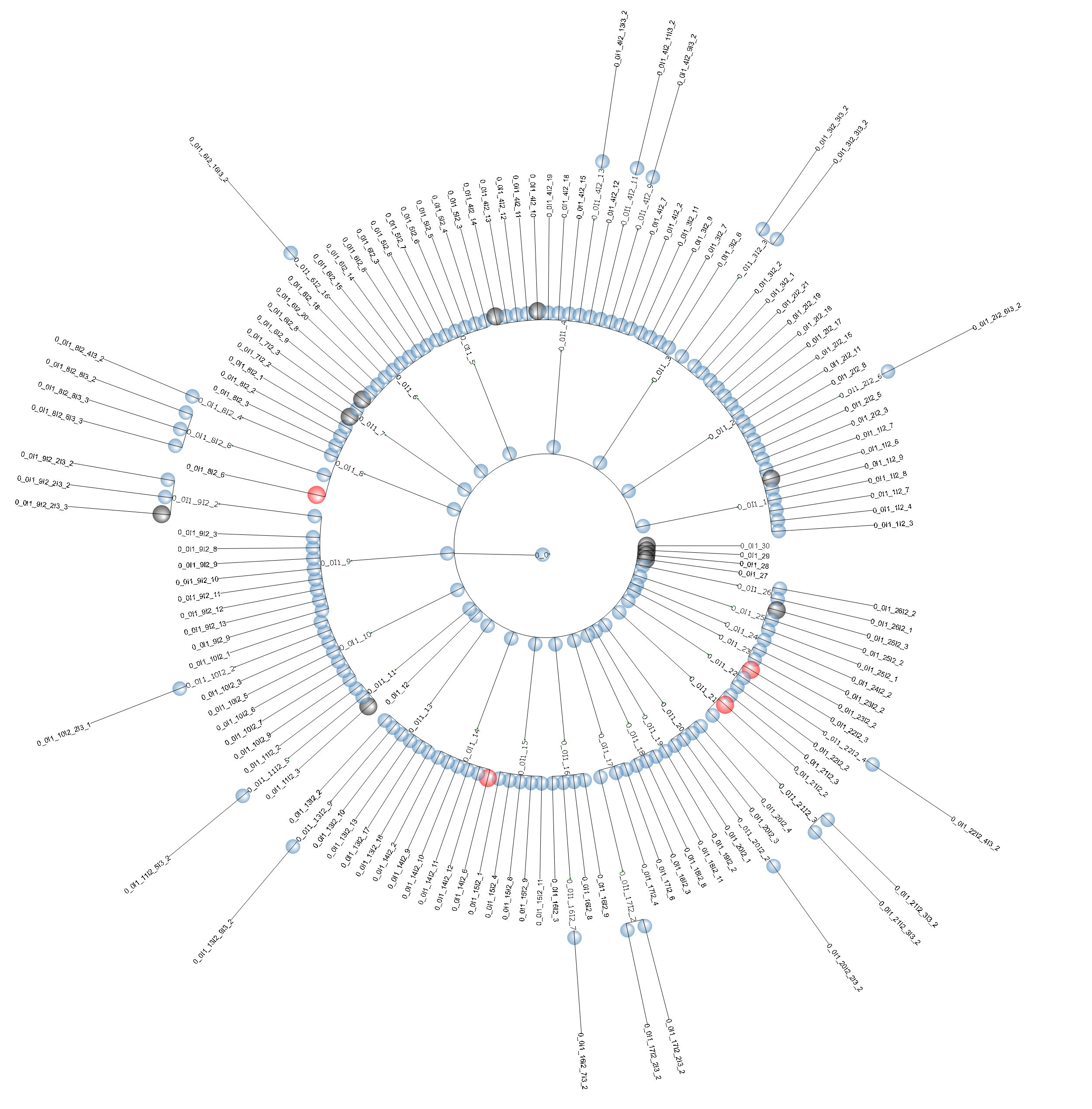}
\caption{Visualization of the full DGT-Net$_L$ as a circular tree after training using YT-VOS18, CDNet, and DAVIS17. The node in the middle is the root of the tree, and the nodes in the first inner circle are children of the root node. The node color defines at which stage the node was added. Blue nodes are added during the initial phase using YT-VOS18. Red nodes are added during the lifelong learning phase using CDNet. Finally, black nodes are added during the few-shot learning phase using DAVIS17. The visualization is made using the ETE toolkit \cite{ete}.}
\label{fig:tree1}
\end{figure}

\textbf{Visualization.} The full DGT after training using the three datasets YT-VOS18, CDNet, and DAVIS17. Fig.~\ref{fig:tree1} visualize DGT-Net$_L$ as a circular tree, starting from the root node in the middle to the leaf nodes at the outermost circle. The written text beside each node is the node name which is formatted as the path of this node from the root node (e.g., a node name $0\_0|1\_3|2\_2$ means the second node in second depth branched from node three in the first depth branched from node zero (root) at depth zero). In the figure, nodes have three colors: blue, red, and black. Blue nodes are the nodes created during training on the first dataset (YT-VOS18). The red nodes are the new nodes created during training on the second dataset (CDNet). The black nodes are those added to DGT during training on the last dataset (DAVIS17). 

\begin{figure*}
\centering
\includegraphics[width=\textwidth]{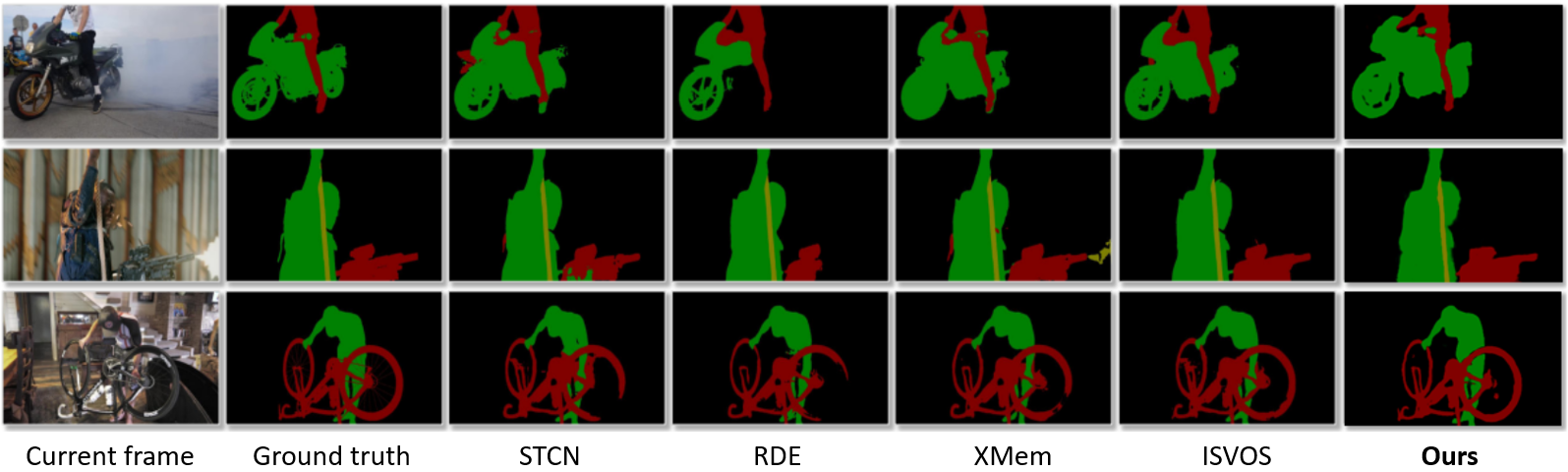}
\caption{Visual results of our proposed model against state-of-the-art video object segmentation models}
\label{fig:visres}
\end{figure*}

\begin{figure}
\hspace{0.2cm}
\includegraphics[width=0.45\textwidth]{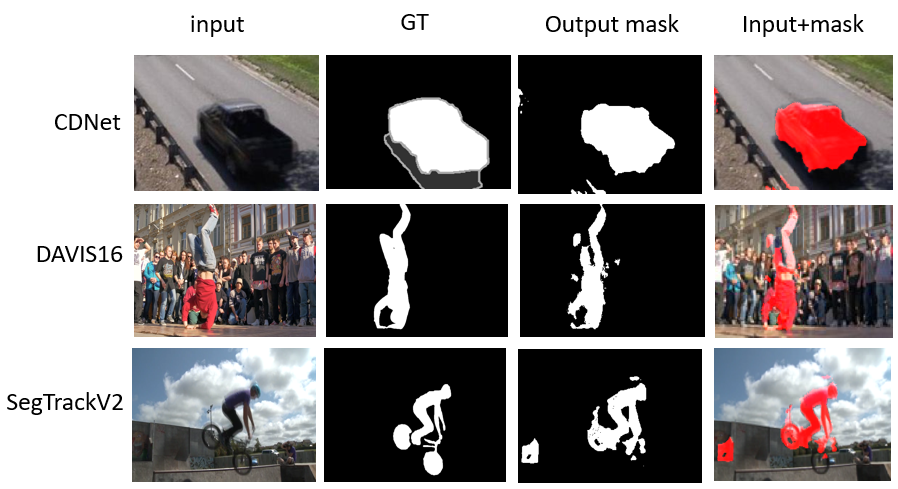}
\caption{Sample failure cases of DGT that occur when the wrong node of the tree is selected.}
\label{fig:failure}
\end{figure}

Fig.~\ref{fig:visres} shows sample results of DGT compared against other models. As shown in the figure, the output of our model is the closest to the ground truth. For example, in the first video, our proposed model was the only model that captured the bike throttle as a part of the bike, while other models detected it as a part of the human. Although the predicted masks in this figure are accurate and almost the same as the ground truth, there are some rare cases where the reference frame used to find a suitable node in the tree is misleading. When a reference frame is misleading, the model might accidentally select the wrong node, causing the model's performance to degrade on all frames of this video. Another case where DGT might fail is when the wrong node is selected due to the greedy selection algorithm of DGT, which might miss the best-performing node in the tree for a given video. The solution for the first problem can be solved by either manually selecting a suitable reference frame for each video or using more than one labeled frame to find the suitable node. The second issue can be solved by traversing all tree nodes instead of using the greedy algorithm to find the most suitable node. However, this will increase the initialization time of DGT for each video. Sample failure cases are shown in Fig.~\ref{fig:failure}.

\subsubsection{\textbf{Few-shot learning for out-of-domain videos}}
A few-shot learning experiment is conducted to further highlight the generalization of DGT. We start with the trained models from the previous experiment (i.e., trained using YT-VOS18, CDNet, and DAVIS16). Then, all models are trained in either 1 or 5 labeled frames (1-shot and 5-shots) from OOD videos (i.e., videos sampled from a dataset not seen during training). After that, the models are evaluated by segmenting the rest of the frames in each video. The dataset used in this experiment is  SegTrackV2. 
As shown in Table~\ref{tb:ffs}, DGT-Net$_L$ outperforms the second-best model by approximately $2.7\%$ in 1-shot and $4\%$ in 5-shot.

\begin{table}[]
\centering
\caption{Few-shot out-of-domain video object segmentation results of DGT compared to state-of-the-art methods on SegTrackV2. The reported numbers are the $\mathcal{F}$-scores for 1-shot (1 labeled frame) and 5-shot (5 labeled frame).}
\begin{tabular}{l|c|c}
\hline
 \multicolumn{1}{l|}{Model}              & \multicolumn{1}{c|}{1-shot}& \multicolumn{1}{c}{5-shot} \\ \hline
XMem \cite{cheng2022xmem}                                                                           & 0.815  & 0.866                                     \\

Net$_S$ (ours)                                                     & 0.727  & 0.780                        \\
Net$_L$ (ours)                                                   & 0.759  & 0.793                        \\
\hline

EWC-Net$_L$ \cite{2cl}                                                        & 0.783 & 0.825                                \\
iCarL-Net$_L$ \cite{5cl}                                                      & 0.790  & 0.830                        \\
ModelZoo-Net$_L$ \cite{1cl}                                                      & 0.833  & 0.875                         \\

TwF-Net$_L$ \cite{3cl}                                                      & 0.784   & 0.827                     \\

CIL-Net$_L$ \cite{4cl}                                                      & 0.793  & 0.835                         \\

\hline
\multicolumn{1}{l|}{\textbf{DGT-Net$_S$ (ours)}}                     & 0.836 & 0.875
\\
\multicolumn{1}{l|}{\textbf{DGT-Net$_L$ (ours)}}                     & \textbf{0.842} & \textbf{0.906}
\\
\hline
\end{tabular}
\label{tb:ffs}
\end{table}
\subsection{DGT analysis and ablation study}

\subsubsection{\textbf{Model size analysis}}
To analyze the proposed model size, we calculated the number of parameters of the model after the last phase in the previous Section.~\ref{sec:eval}. Our model is divided into the tree (DGT) and the task-specific network (Net$_S$ or Net$_L$). We report the number of parameters in each part separately. In Table~\ref{tb:size}, the number of parameters of the full tree is reported as "Full model (Param. $\#$)". While the number of parameters of the task-specific network is reported as an "Inference model (Param. $\#$)". The values in the table are in millions of parameters. The number of parameters for state-of-the-art models is the same in the full model and inference model. As shown in Table~\ref{tb:size}, DGT-Net$_S$ started with $7.7$ million parameters and, during training, expanded to $186.1$ million parameters. On the other hand, the DGT-Net$_L$ base network is $10.3$ million parameters, and the full tree size is $354.6$ million parameters.

\begin{table}[]
\centering
\caption{Model size analysis of proposed work against other models. The number of parameters is in millions.}
\begin{tabular}{l|c|c}
\hline
 \multirow{2}*{Model}             & \multirow{2}*{\shortstack[c|]{Full model\\(Param. $\#$)}}& \multirow{2}*{\shortstack[c|]{Inference model\\(Param. $\#$)}} \\& &  \\\hline

XMem \cite{cheng2022xmem}                                                                           & 62.1  & 62.1                                        \\

ModelZoo-Net$_L$ \cite{ramesh2021model}                                                    & 594  & 594                         \\
\hline
\multicolumn{1}{l|}{\textbf{Net$_S$ (ours)}}                     & 7.7  & 7.7                 \\
\multicolumn{1}{l|}{\textbf{DGT-Net$_S$ (ours)}}                     & 186.1   &  7.7
\\ 
\multicolumn{1}{l|}{\textbf{Net$_L$ (ours)}}                     & 10.3 &  10.3                   \\
\multicolumn{1}{l|}{\textbf{DGT-Net$_L$ (ours)}}                     & 354.6   & 10.3                 \\\hline
\end{tabular}
\label{tb:size}
\end{table}

\begin{figure}
\centering
\includegraphics[width=0.45\textwidth]{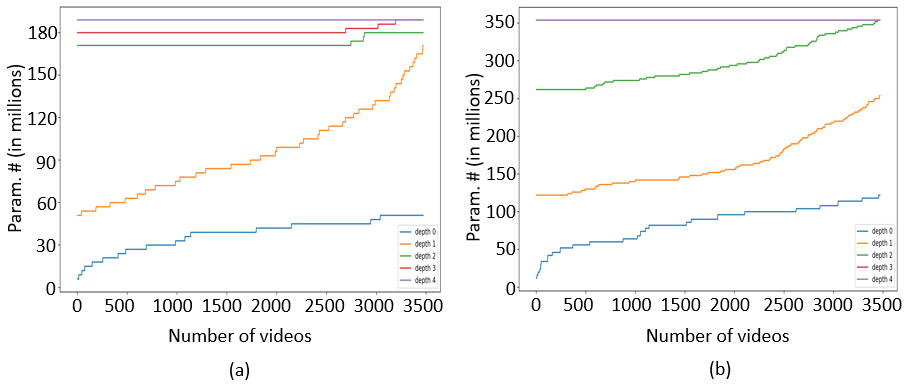}
\caption{The increase in the number of parameters of the model versus the number of videos during each depth in the "Building base tree" phase of DGT using the YT-VOS18 dataset. (a) shows the increase in the number of parameters of DGT-Net$_S$. While (b) shows the increase in the number of parameters of DGT-Net$_L$.}
\label{fig:growth}
\end{figure}

In Fig.~\ref{fig:growth}, The growth in the number of parameters of DGT against the number of videos is shown in different depths of the "Building Base Tree" training phase. As shown in Fig.~\ref{fig:growth}, the model size increases noticeably during the training at depths $1$ and $2$ for both models DGT-Net$_S$ and DGT-Net$_L$. Then, the model size increases slightly at depth $3$. On the other hand, at depths $4$ and $5$, the model size is almost the same (i.e., the model gets saturated after depth 3). We calculated the average growth rate of model size as the total number of parameters added (i.e., full tree size minus base model size) divided by the number of videos. The number of training videos in YT-VOS18 is $3,471$. The growth rate for DGT-Net$_S$ and DGT-Net$_L$ is $51,397$ and $99,279$ parameters, respectively.

\subsubsection{\textbf{Runtime analysis}}
Our proposed model runtime is analyzed and compared with the state-of-the-art model. We split the runtime into initialization time and inference time, as our model has two phases during testing: task-specific network generation and inference. The initialization time is the time taken to generate a task-specific network, which is done only once per video. The inference time is the time taken to produce the output of an input frame.
Finally, frame per second (FPS) is reported to show how many frames can be processed by our model within the same video. The reported FPS is calculated using a single A6000 GPU. As shown in Table~\ref{tb:runtime}, DGT-Net$_S$ is much faster than XMem in inference time. As shown in the table, DGT-Net$_S$ can process $38.4$ frames per second which is $16.2$ FPS faster than XMem. This processing speed is due to the small size of the network used during inference (7.7M parameters only), where XMem size during inference is (M parameters). When DGT is used with Net$_L$, the FPS degrades slightly to $32.2$ FPS at the expense of the performance gained of $2.7\%$ in average $\mathcal{F}$ over all three datasets, as shown in Table.\ref{tb:gfs}. However, the initialization time for both DGT-Net$_S$ and DGT-Net$_L$ are approximately the same. This is because initialization time is affected by the number of nodes in DGT, while inference time is affected by the number of parameters in the network. On the other hand, ModelZoo has a huge inference time because each frame must be processed by the entire model, as the output is the average of all sub-networks.

\begin{table}[]
\centering
\caption{Runtime analysis of proposed work against other models. The reported runtime is in seconds.}
\begin{tabular}{l|c|c|c}
\hline
 \multirow{2}*{Model}             & \multirow{2}*{\shortstack[c|]{Initialization time\\(per video)}}& \multirow{2}*{\shortstack[c|]{Inference time\\(per frame)}}& \multirow{2}*{FPS} \\& & & \\\hline

XMem \cite{cheng2022xmem}                                                                           & 0.0  & 0.044 & 22.6                                       \\

ModelZoo-Net$_L$ \cite{ramesh2021model}                                                    & 0.0  & 1.631 & 0.6                        \\
\hline
\multicolumn{1}{l|}{\textbf{Net$_S$ (ours)}}                     & 0.00 & 0.026 & \textbf{38.4}                  \\

\multicolumn{1}{l|}{\textbf{DGT-Net$_{S}$ (ours)}}                     & 3.55 & 0.026 &  \textbf{38.4}
\\ 
\multicolumn{1}{l|}{\textbf{Net$_L$ (ours)}}                     & 0.00 & 0.031 & 32.2                  \\

\multicolumn{1}{l|}{\textbf{DGT-Net$_{L}$ (ours)}}                     & 4.17 & 0.031 & 32.2                   \\\hline
\end{tabular}
\label{tb:runtime}
\end{table}

\section{Conclusion}
This paper presents DGT, a video object segmentation technique using a dynamically growing tree of sub-networks. DGT can dynamically grow by adding sub-networks to its tree structure as new videos are added to the training set. DGT's novel architecture and novel learning technique lead to effective lifelong learning with minimal catastrophic forgetting of previously trained videos. DGT outperforms state-of-the-art video object segmentation models in single and multiple-source in-domain experiments. Moreover, DGT can effectively perform well on out-of-domain videos with a few labeled frames. The proposed work is evaluated using three experiments: single-source domain, multiple-source domains, and few-shot out-of-domain. The results in the first experiment (single-source in-domain) show that DGT outperforms state-of-the-art models with respect to 
$\mathcal{F}$-score by $0.2\%$ and $3.5\%$ in DAVIS16 and DAVIS17, respectively. Moreover, in the second experiment (multi-sources in-domain), the results show superior performance compared to state-of-the-art video object segmentation and other lifelong learning techniques with an average performance increase in the F-score of $6.9\%$ with minimal catastrophic forgetting. Finally, in the last experiment (few-shot out-of-domain), the performance of DGT is higher than the state-of-the-art video object segmentation by $2.7\%$ and $4\%$ in 1-shot and 5-shot, respectively.


\end{document}